\documentclass[10pt,twocolumn,letterpaper]{article}
\usepackage[pagenumbers]{cvpr} 

\definecolor{cvprblue}{rgb}{0.21,0.49,0.74}
\usepackage[pagebackref,breaklinks,colorlinks,allcolors=cvprblue]{hyperref}

\usepackage{amsmath,amsfonts,bm}

\def\eqref#1{equation~\ref{#1}}

\def\1{\bm{1}}

\DeclareMathAlphabet{\mathsfit}{\encodingdefault}{\sfdefault}{m}{sl}
\SetMathAlphabet{\mathsfit}{bold}{\encodingdefault}{\sfdefault}{bx}{n}

\usepackage{hyperref}
\usepackage{url}
\usepackage{booktabs}   
\usepackage{multirow}   
\usepackage{subcaption}
\usepackage{wrapfig}
\usepackage{makecell}
\usepackage{xspace}
\usepackage{enumitem}
\usepackage{amsmath}
\usepackage{listings}
\usepackage{algorithm}
\usepackage{algorithmic}
\usepackage{amsfonts,amssymb}
\usepackage{mathrsfs}
\usepackage{subcaption}
\usepackage{natbib}
\setcitestyle{numbers,square}
\usepackage{graphicx}       
\usepackage{minitoc}
\usepackage[table]{xcolor}
\newcommand{\hhide}[1]{}

\title{Video Generation Models Are Good Latent Reward Models}

\author{
Xiaoyue Mi$^{1,\dagger *}$ \quad 
Wenqing Yu$^{2,*}$ \quad
Jiesong Lian$^{3}$ \quad
Shibo Jie$^{4}$ \quad 
Ruizhe Zhong$^{5}$ \quad \\
Zijun Liu$^{6}$ \quad
Guozhen Zhang$^{7}$ \quad
Zixiang Zhou$^{2}$ \quad
Zhiyong Xu$^{2}$ \quad \\
Yuan Zhou$^{2,\ddagger}$ \quad 
Qinglin Lu$^{2}$ \quad
Fan Tang$^{1,\S}$\\
$^1$University of Chinese Academy of Sciences \quad
$^2$Tencent Hunyuan \quad \\
$^3$Huazhong University of Science and Technology \quad  \\
$^4$Peking University \quad 
$^5$Shanghai Jiao Tong University \quad
$^6$Tsinghua University \quad
$^7$Nanjing University \\
\vspace{1mm}
}
\begin{document}

\twocolumn[{
\maketitle

\begin{center}
  \vspace{-8mm}
  \label{fig:head}
  \includegraphics[width=1\textwidth]{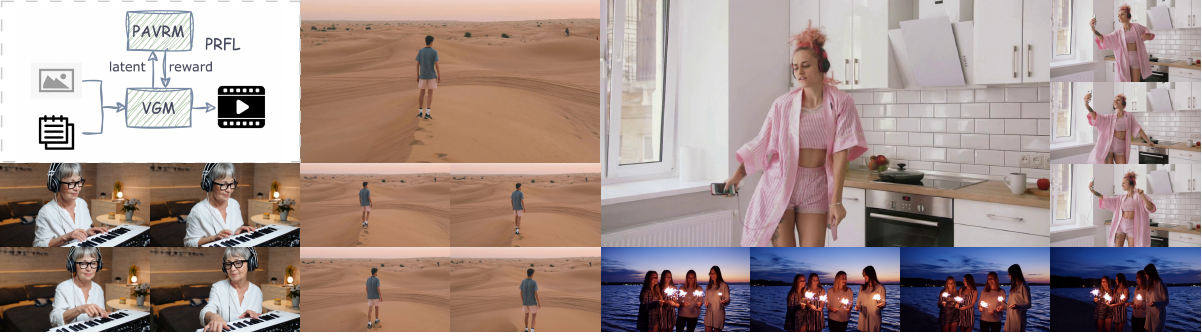}
  \captionof{figure}{PRFL generates high-quality videos with enhanced motion quality. We show results on image-to-video (frames of different sizes) and text-to-video (frames of same size) generation tasks, showing the representative frames from each video.}
  \label{fig:head}
\end{center}
}]
\footnotetext[1]{$\dagger$ Work done during internship at Tencent Hunyuan.}
\footnotetext[2]{$*$ Equal contribution. mxysdu@gmail.com}
\footnotetext[3]{$\ddagger$ Project leader.}
\footnotetext[4]{$\S$ Corresponding author. tfan.108@gmail.com}

\begin{abstract}
Reward feedback learning (ReFL) has proven effective for aligning image generation with human preferences.
However, its extension to video generation faces significant challenges.
Existing video reward models rely on vision-language models designed for pixel-space inputs, confining ReFL optimization to near-complete denoising steps after computationally expensive VAE decoding.
This pixel-space approach incurs substantial memory overhead and increased training time, and its late-stage optimization lacks early-stage supervision, refining only visual quality rather than fundamental motion dynamics and structural coherence.
In this work, we show that pre-trained video generation models are naturally suited for reward modeling in the noisy latent space, as they are explicitly designed to process noisy latent representations at arbitrary timesteps and inherently preserve temporal information through their sequential modeling capabilities. Accordingly, we propose Process Reward Feedback Learning~(PRFL), a framework that conducts preference optimization entirely in latent space, enabling efficient gradient backpropagation throughout the full denoising chain without VAE decoding.
Extensive experiments demonstrate that PRFL significantly improves alignment with human preferences, while achieving substantial reductions in memory consumption and training time compared to RGB ReFL.
\end{abstract}
\section{Introduction}
\label{sec:introduction}
Recent video generation models~\cite{lin2024open,kong2024hunyuanvideo,wan2025,gao2025seedance} have demonstrated remarkable visual fidelity, producing photorealistic content across diverse applications~\cite{liu2024video,polyak2024movie,henschel2025streamingt2v}.
Nevertheless, aligning these models with complex human preferences, particularly concerning motion quality~\cite{liu2025improving}, physical plausibility~\cite{bansal2025videophy}, and prompt following~\cite{Cheng_2025_ICCV}, remains a fundamental challenge~\cite{zeng2024dawn}.
Reward feedback learning (ReFL)~\cite{xu2023imagereward,prabhudesai2023aligning,clarkdirectly2024} has emerged as a promising way for preference alignment in image generation.
Nevertheless, its direct application to video generation encounters critical computational and optimization barriers that fundamentally limit its applicability.

\begin{figure}
    \centering
    \includegraphics[width=1\linewidth]{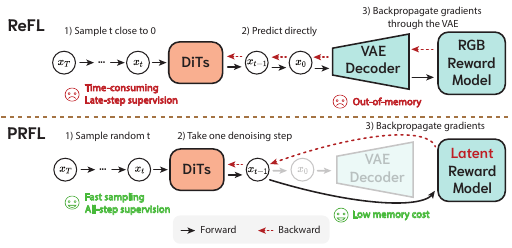}
    \vspace{-8mm}
\caption{Comparison between RGB ReFL and our PRFL. RGB reward models require near-complete denoising and VAE decoding to RGB space, introducing \textit{evaluation delay}, \textit{GPU memory bottleneck}, and \textit{insufficient supervision} of early denoising stages where structure and motion are formed. PRFL eliminates these limitations by performing reward modeling directly in latent space with timestep-aware training.}
    \label{fig:motivation}
    \vspace{-6mm}
\end{figure}
\begin{figure*}
    \centering
    \includegraphics[width=0.8\linewidth]{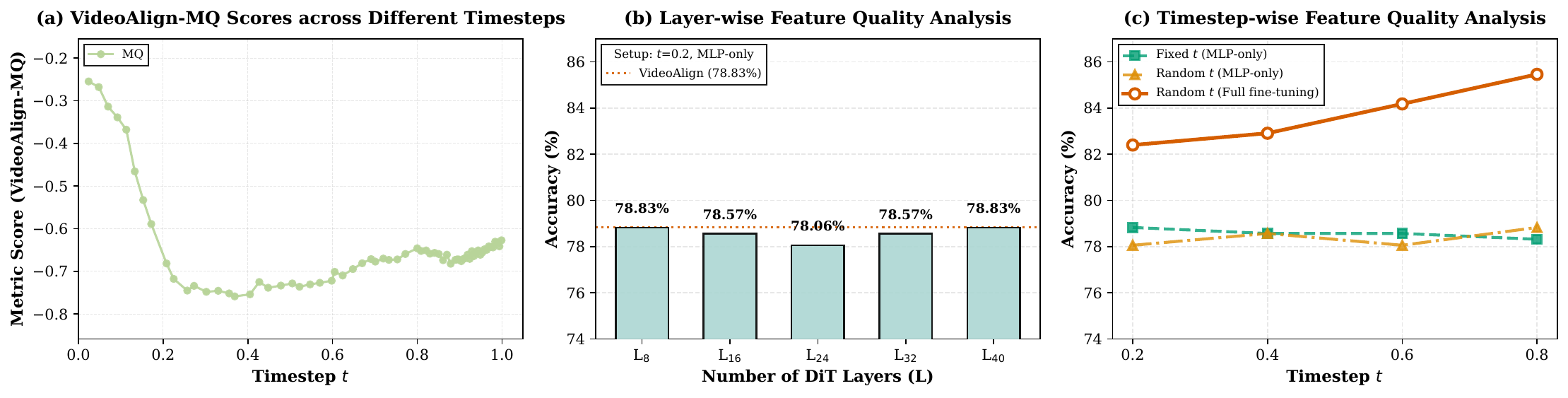}
    \vspace{-4mm}
    \caption{Analysis on VGM Features. (a) VLM-based reward model (VideoAlign-MQ) exhibits poor timestep generalization with fluctuating scores. (b) VGM features from any DiT layer uniformly achieve 78.8\% accuracy, matching VLM baseline. (c) Timestep-aware fine-tuning unlocks VGM's full potential, achieving 85.46\% accuracy with peak performance at early timesteps (t=0.8).}
    \label{fig:analysis}
    \vspace{-4mm}
\end{figure*}
Typical ReFL approaches rely on outcome-based reward models built upon Vision-Language Models (VLMs)~\cite{unifiedreward,xu2023imagereward,he2024videoscore,liu2025improving,he2025videoscore2}.
As shown in Fig.~\ref{fig:motivation}, such outcome-based reward models focus on videos in RGB space, which imposes three interconnected limitations due to their inherent post-hoc nature for diffusion-based video generation models (VGMs).
\textbf{Initially}, the RGB inputs for VLMs require near-complete denoising, which introduces significant \textit{evaluation delay} and drastically slows down training iterations.
\textbf{Furthermore}, this necessity leads to a \textit{GPU memory bottleneck}, as backpropagation through the VAE decoder for all video frames frequently causes GPU memory overflow.
\textbf{Ultimately}, the issue of \textit{insufficient supervision} occurs because rewards are only applied at the final steps, failing to directly guide early generation stages when structure and motion are formed.
Although some recent methods~\cite{wu2024deep,Guo_2025_ICCV} attempt to distribute gradient updates via gradient stopping or trajectory shortcuts, they still depend on fully denoised frames and costly VAE decoding for reward model construction.
ContentV~\cite{lin2025contentv} bypasses VAE decoding by optimizing only the first frame, yet sacrifices holistic video quality assessment, while DOLLAR~\cite{ding2025dollar} employs a VLM-based latent reward model but lacks fine-grained timestep-wise supervision.

Alternatively, VGMs themselves have shown potential as reward sources, offering process-level supervision throughout the denoising trajectory. 
For instance, LPO~\cite{zhang2025diffusion} pioneered using diffusion models as noise-aware latent reward models for image generation, while VideoAlign~\cite{liu2025improving} briefly mentioned using video generation models for inference-time guidance.
However, the use of video generation models as train-time reward models remains largely unexplored.
We posit that VGMs possess unique properties for reward modeling: (1) inherent noise-aware feature extraction at arbitrary denoising step, (2) sensitivity to generation artifacts, and (3) native support for full-sequence processing without frame sampling.
Through experimental analysis, we verify that the latent features of VGMs are suitable for determining the quality of videos.
Yet a critical question remains: \textbf{how to repurpose VGMs into effective reward models that compress comprehensive spatiotemporal features while maintaining noise-level sensitivity}.

In this study, we tackle these challenges through process-level reward modeling and optimization, enlarging VGMs.
First, we propose a {Process-Aware Video Reward Model (PAVRM)}, which repurposes video generation models to evaluate quality directly from noisy latent representations at arbitrary timesteps.
PAVRM employs learnable query vectors to compress variable-length spatiotemporal features into a compact video quality-aware token, inherently encouraging the model to learn quality-relevant patterns rather than memorizing content-specific correlations.
Furthermore, we develop Process Reward Feedback Learning (PRFL) to optimize generation quality.
During training, PRFL randomly samples timesteps and maximizes process rewards through a single gradient, with no VAE decoding overhead and the full-trajectory learning signal distribution.
Extensive experiments demonstrate that PRFL achieves \textbf{substantial improvements} (up to \textbf{+56.00} in dynamic degree, \textbf{+21.52} in human anatomy) with \textbf{significant memory savings} and at least \textbf{1.4$\times$ faster training} over RGB ReFL.
Our contributions can be summarized as follows:
\begin{itemize}
\item We propose a \textbf{process-aware video reward model (PAVRM)}, which employs query-based aggregation to efficiently handle variable-length videos with timestep sensitivity, and keep artifact awareness throughout the denoising process based on pre-trained video models.
\item We introduce \textbf{process reward feedback learning (PRFL)}, an efficient video post-training framework that operates in latent space by sampling random timesteps and optimizing through single denoising steps, without VAE decoding and distributing reward across the full denoising process.
\item Experiments show our approach improves motion quality while saving substantial GPU memory and at least accelerating training by 1.4$\times$ compared to RGB ReFL.
\end{itemize}

\section{Related Work}
\subsection{Visual Reward Feedback Learning}
Reward feedback learning has proven effective for improving visual generation quality. 
Early works~\cite{xu2023imagereward,clarkdirectly2024,prabhudesai2023aligning} finetune diffusion models using differentiable reward signals from human preferences, but face a depth-efficiency dilemma: step-by-step backpropagation through extensive denoising steps incurs prohibitive memory costs, while training only final steps fails to optimize low-level objectives like symmetry. Recent approaches~\cite{wu2024deep,Guo_2025_ICCV} mitigate this through gradient stopping or trajectory-preserving shortcuts.
However, these image-focused methods encounter severe computational barriers in video generation due to multi-frame synthesis and gradient-enabled VAE decoding overhead. ContentV~\cite{lin2025contentv} optimizes only the first frame but cannot capture comprehensive video quality.
Critically, almost existing approaches require pixel-space rewards, necessitating VAE decoding.
While DOLLAR~\cite{ding2025dollar} leverages a VLM as a latent reward model for distillation tasks to avoid costly VAE decoding, it lacks timestep-wise optimization capability.
In this paper, we propose a timestep-wise latent reward model operating directly in latent space, enabling gradient backpropagation at arbitrary timesteps while maintaining computational tractability.

\subsection{Reward Models for Video Generation}
Reward models for RLHF~\cite{ouyang2022training} are typically trained on human preferences~\cite{sun2025rethinking} and categorized into outcome models~\cite{shao2024deepseekmath} supervising final results and process models~\cite{lightman2024lets,wang2024math,zhang2025lessons} supervising intermediate steps.
Current video reward models focus on outcome models.
Early video reward models adapted image-based approaches~\cite{yuan2024instructvideo} or introduced VLMs for quality assessment~\cite{wu2024boosting,liu2025improving,he2024videoscore,he2025videoscore2,mougradeo}.
However, these outcome-based reward models operate only on near-complete outputs, lacking direct guidance for the critical early denoising stages where fundamental video attributes are established.
However, these outcome-based reward models need almost fully denoising stages and lack direct guidance to early denoising stages. 
LPO~\cite{zhang2025diffusion} pioneered using diffusion models as noise-aware latent reward models for image generation, a promising process reward attempt.
However, different from image generation, video generation introduces substantially greater complexity.
VideoAlign~\cite{liu2025improving} briefly mentioned using VGMs as guidance during inference augmentation, while using VGMs as reward models during training remained unexplored.
We comprehensively investigate VGMs as latent reward models, offering frame-continuous processing, artifact sensitivity, and timestep-aware rewards aligned with the generation process, with effective integration into the ReFL and other online reinforcement learning methods.

\begin{figure*}
    \centering
    \includegraphics[width=1.\linewidth]{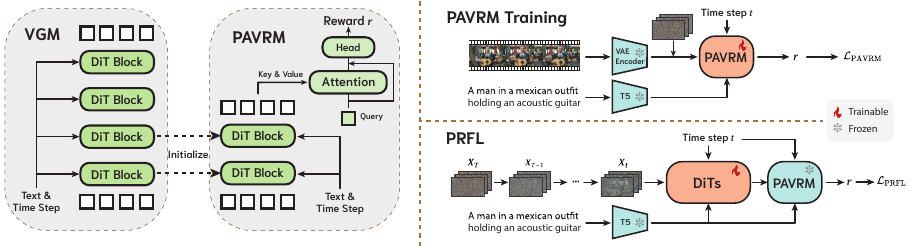}
    \caption{Overview of our process-aware video generation alignment framework. Left: Architecture of the Video Generation Model (VGM) and Process-Aware Video Reward Model (PAVRM). Right: Two-stage training pipeline. PAVRM training with reward prediction from noisy latents. Process Reward Feedback Learning (PRFL) optimizing VGM through latent-space reinforcement learning at randomly sampled timesteps.}
    \label{fig:pipeline}
\end{figure*}
\section{Preliminaries and Feasibility Analysis}
\label{sec:pre}
\paragraph{Rectified Flow.}
Recently, video generation models~\cite{kong2024hunyuanvideo,wan2025,gao2025seedance} operate in rectified flow~\cite{lipman2023flow,liu2023flow}, which establishes a continuous transport between data and noise distributions. Specifically, given clean data $\mathbf{x}_0$ sampled from distribution $q(\mathbf{x}_0)$ and noise $\mathbf{x}_1$ from distribution $p(\mathbf{x}_1)$, the framework defines a time-dependent interpolation path:
\begin{equation}
    \mathbf{x}_t = (1-t)\mathbf{x}_0 + t\mathbf{x}_1, \quad t \in [0,1].
    \label{eq:flow_forward}
\end{equation}
A neural network (usually several DiT blocks) parameterized by $\theta$ predicts the velocity field $\mathbf{v}_\theta(\mathbf{x}_t, t)$ at each timestep, trained via the flow matching objective, which also as supervised fine-tuning~(SFT) loss:
\begin{equation}
    \mathcal{L}_{\text{FM}}(\theta) = \mathbb{E}_{t \sim \mathcal{U}(0,1),\;\mathbf{x}_0\sim q(\mathbf{x}_0),\;\mathbf{x}_1\sim p(\mathbf{x}_1)} \bigl[\|\mathbf{v}_\theta(\mathbf{x}_t, t) - \mathbf{v}\|^2\bigr],
    \label{eq:flow_loss}
\end{equation}
where $\mathbf{v} = \mathbf{x}_1 - \mathbf{x}_0$ denotes the data-to-noise transport direction.

\paragraph{Reward Feedback Learning.}
Reward Feedback Learning~(ReFL)~\cite{xu2023imagereward} optimizes diffusion models by backpropagating reward signals through the denoising process. Given a reward model $r_\phi$ and generated output $\mathbf{x}_0$, ReFL samples a timestep $t$ from the late denoising stage and computes gradients from the reward score.
To prevent reward over-optimization, the objective combines reward-based loss with SFT regularization:
\begin{equation}
\label{eq:refl_loss}
\mathcal{L}_{\text{ReFL}} = -\lambda \mathbb{E}_{\mathbf{x}_0 \sim \text{VGM}_\theta} \left[ r_\phi(\mathcal{D}(\mathbf{x}_0)) \right] + \mathcal{L}_{\text{FM}}(\theta),
\end{equation}
where $\mathcal{D}$ denotes the VAE decoder.

\paragraph{Feasibility Analysis of VGM-based Reward Models.}
We analyze whether VGMs can serve as effective reward models using motion quality as a case study. 

\noindent \textit{VLM fails on noisy inputs (Fig.~\ref{fig:analysis}(a)).} 
VideoAlign~\cite{liu2025improving}, a representative VLM-based video reward model trained on large-scale preference data, exhibits severe performance degradation across different denoising timesteps. When evaluating latents decoded to RGB space, the logit scores fluctuate dramatically between early and late timesteps, revealing poor generalization to high-noise regions.

\noindent \textit{VGM features encode motion efficiently (Fig.~\ref{fig:analysis}(b)).}
We employ a simple MLP as a linear probe to assess VGM feature quality. Remarkably, probing features from any DiT layer ($\mathrm{L}_{8}$ to $\mathrm{L}_{40}$) achieves uniform performance at 78.8\%, matching the VideoAlign baseline. This confirms that motion dynamics are efficiently distributed throughout the network, enabling the use of only the first 8 layers for a more efficient reward model.

\noindent \textit{Timestep-aware fine-tuning unlocks VGM's potential (Fig.~\ref{fig:analysis}(c)).}
While MLP-only probing (both fixed and random timestep) fails to surpass the VLM baseline, coupling full fine-tuning with random timestep sampling yields a dramatic leap to \textbf{85.46\%} ($\sim$\textbf{6.6\%} absolute gain), particularly when sampling features from earlier noisy states (t=0.8). This validates that fine-tuned VGMs provide superior, timestep-robust signals for learning motion quality rewards.

\begin{algorithm}[t]
\small
\caption{Process Reward Feedback Learning (PRFL)}
\begin{algorithmic}[1]
\STATE \textbf{Input:} Velocity predictor $\mathbf{v}_\theta$, PAVRM $F_\phi$, VAE encoder $\mathcal{E}$, datasets $\mathcal{D}_{\text{PRFL}}$, $\mathcal{D}_{\text{SFT}}$, reward weight $\lambda$, total timesteps $T$, timestep interval $\Delta t$, learning rate $\eta$
\FOR{each training iteration}
    \STATE \textcolor{gray}{// Supervised Fine-Tuning Regularization}
    \STATE Sample $(V, I, p) \sim \mathcal{D}_{\text{SFT}}$ and $t_{\text{sft}} \sim \mathcal{U}(0, 1)$
    \STATE Compute $\mathbf{x}_0 = \mathcal{E}(V)$, sample $\mathbf{x}_1 \sim \mathcal{N}(0, \mathbf{I})$
    \STATE Construct $\mathbf{x}_{t_{\text{sft}}} = (1 - t_{\text{sft}}) \mathbf{x}_0 + t_{\text{sft}} \mathbf{x}_1$
    \STATE Compute $\mathcal{L}_{\text{FM}} = \|\mathbf{v}_\theta(\mathbf{x}_{t_{\text{sft}}}, t_{\text{sft}}) - (\mathbf{x}_1 - \mathbf{x}_0)\|_2^2$
    \STATE Update $\theta \leftarrow \theta - \eta \nabla_\theta \mathcal{L}_{\text{FM}}$
    \STATE \textcolor{gray}{// Process Reward Feedback Learning}
    \STATE Sample $(I, p) \sim \mathcal{D}_{\text{PRFL}}$, $\mathbf{x}_T \sim \mathcal{N}(0, \mathbf{I})$, and $t \sim \mathcal{U}(0, T-1)$
    \FOR{$j = T, T-1, \ldots, t+1$}
        \STATE \textbf{no grad:} $\mathbf{x}_{j-1} \leftarrow \mathbf{x}_j - \Delta t \cdot \mathbf{v}_\theta(\mathbf{x}_j, j)$
    \ENDFOR
    \STATE \textbf{with grad:} $\mathbf{x}_t \leftarrow \mathbf{x}_{t+1} - \Delta t \cdot \mathbf{v}_\theta(\mathbf{x}_{t+1}, t+1)$
    \STATE Compute $r_t = F_\phi(\mathbf{x}_t, t, I, p)$ and $\mathcal{L}_{\text{PRFL}} = -\lambda \cdot r_t$
    \STATE Update $\theta \leftarrow \theta - \eta \nabla_\theta \mathcal{L}_{\text{PRFL}}$
\ENDFOR
\end{algorithmic}  
\label{alg:prfl}

\end{algorithm}
\vspace{-3mm}
\section{Method}
\vspace{-2mm}
\subsection{Overview}
\label{sec:overview}
We propose process reward feedback learning, a new ReFL framework in the video generation task to address GPU memory and training time limitations.
As shown in Fig.~\ref{fig:pipeline}, the full process consists of two stages: (1) training a Process-Aware Video Reward Model (PAVRM, Sec.~\ref{sec:pavrm}) adapted from the VGM that evaluates video quality directly from noisy latents at arbitrary diffusion timesteps, and (2) optimizing the VGM using Process Reward Feedback Learning (PRFL, Sec.~\ref{sec:prfl}).

\vspace{-2mm}
\subsection{Process-Aware Video Reward Model}
\label{sec:pavrm}

\paragraph{Architecture of PAVRM.} 
PAVRM leverages features from a frozen pre-trained VGM to predict preference scores directly from noisy latents, as illustrated in Fig.~\ref{fig:pipeline} (left).
Given a noisy latent video $\mathbf{x}_t \in \mathbb{R}^{F \times H \times W \times C}$ at timestep $t$ and a text prompt $p$, we use the first eight DiT blocks of the backbone as \textbf{spatiotemporal feature-extractor}:
\begin{equation}
    \mathbf{h} = \text{DiT}_\phi(\mathbf{x}_t, t, \mathcal{T}(p)) \in \mathbb{R}^{F \times H \times W \times D},
\end{equation}
where $\mathcal{T}(\cdot)$ denotes the text encoder, and $D$ denotes the feature dimension. These intermediate representations naturally encode motion quality signals at various noise levels, as demonstrated in Fig.~\ref{fig:analysis}.

To handle variable-length video features and extract discriminative representations, we employ a query attention as \textbf{query-based spatiotemporal aggregation} that adaptively compresses spatiotemporal information into a fixed-size embedding. Specifically, we flatten the spatiotemporal features $\mathbf{h} \in \mathbb{R}^{F \times H \times W \times D}$ into $\hat{\mathbf{h}} \in \mathbb{R}^{N \times D}$ where $N = F \cdot H \cdot W$, and compute:
\begin{equation}
    \mathbf{z}_{\text{obs}} = \frac{\exp(\mathbf{q} (\hat{\mathbf{h}} \mathbf{W}_K)^T / \sqrt{D})}{\sum \exp(\mathbf{q} (\hat{\mathbf{h}} \mathbf{W}_K)^T / \sqrt{D})} (\hat{\mathbf{h}} \mathbf{W}_V) \in \mathbb{R}^{1 \times D},
\end{equation}
where $\mathbf{q} \in \mathbb{R}^{1 \times D}$ is a learnable query vector, and $\mathbf{W}_K, \mathbf{W}_V \in \mathbb{R}^{D \times D}$ are projection matrices. The final representation $\mathbf{z} = \mathbf{z}_{\text{obs}}+\mathbf{q} \in \mathbb{R}^{D}$ concatenates observation features with content-agnostic quality priors, enabling the model to reason about generation quality independent of content correlations.
A three-layer MLP as \textbf{head layer} maps $\mathbf{z}$ to the reward score $r_\phi(\mathbf{x}_t, t, p) \in \mathbb{R}$.

\vspace{-4mm}
\paragraph{Training Objective.} 
PAVRM is trained on a binary preference dataset $\mathcal{D}_{\text{RM}} = \{(V_i, p_i, y_i)\}_{i=1}^N$ where $y_i \in \{0, 1\}$ indicates motion quality satisfaction. Following the rectified flow formulation (Eq.~(\ref{eq:flow_forward})), we construct noisy latents $\mathbf{x}_t = (1-t)\mathbf{x}_0 + t \mathbf{x}_1$ with $\mathbf{x}_1 \sim \mathcal{N}(0, \mathbf{I})$ and $t \sim \mathcal{U}(0, 1)$, where $\mathbf{x}_0 = \mathcal{E}(V)$. The training objective minimizes:
\begin{equation}
\begin{split}
    \mathcal{L}_{\text{PAVRM}} = & -\mathbb{E}_{t, (V,p,y)} \bigg[ y \log \sigma(r_\phi(\mathbf{x}_t, t, p)) \\
    & + (1-y) \log (1-\sigma(r_\phi(\mathbf{x}_t, t, p)))\bigg],
\end{split}
    \label{eq:loss_pavrm}
\end{equation}
where $t \sim \mathcal{U}(0,1)$ and $(V,p,y) \sim \mathcal{D}_{\text{RM}}$. We freeze the VAE encoder $\mathcal{E}$ while jointly optimizing the DiT blocks, query vector $\mathbf{q}$, and MLP parameters.

A critical design of PAVRM is the random sampling of timesteps $t \in [0,1]$ during training. Unlike outcome-based reward models that only evaluate clean outputs ($t \approx 0$), PAVRM learns stage-appropriate quality assessment across the entire denoising trajectory. This is justified in rectified flow: since $\mathbf{x}_t$ follows a deterministic linear interpolation from noise to data, preference labels of final outputs naturally propagate to intermediate states. 
This process-aware training enables effective guidance throughout the generation process, as demonstrated in Sec.~\ref{sec:exp}.

\subsection{Process Reward Feedback Learning}
\label{sec:prfl}
As discussed in Sec.~\ref{sec:pre}, vanilla RGB ReFL (Eq.~(\ref{eq:refl_loss})) optimizes only final denoised outputs $\mathbf{x}_0$, requiring VAE decoding $\mathcal{D}(\mathbf{x}_0)$ and pixel-space reward computation—prohibitively expensive for high-resolution videos. Moreover, sampling timesteps only from late denoising stages fails to improve early-stage motion planning, which critically determines overall physical coherence. PRFL addresses these limitations by performing reward optimization at arbitrary intermediate timesteps directly in latent space, eliminating VAE decoding while distributing learning signals across the entire generation trajectory.
\vspace{-4mm}
\paragraph{Training Procedure.}
As shown in Alg.~\ref{alg:prfl}, we extend the ReFL framework by replacing outcome rewards with process-level rewards from PAVRM.
Given the text prompt $p$ from the dataset $\mathcal{D}_{\text{PRFL}}$, we sample the initial noise $\mathbf{x}_1 \sim \mathcal{N}(0, \mathbf{I})$ and the target timestep $s \sim \mathcal{U}(0, 1)$.
In practice, we perform gradient-free denoising rollouts from $t=1$ to $t=s+\Delta t$, and then execute one gradient-enabled step:
\begin{align}
    & \mathbf{x}_{t-\Delta t} = \mathbf{x}_t - \Delta t \cdot \mathbf{v}_\theta(\mathbf{x}_t, t, p)  \text{(w./o. grad)}, \\
    & \mathbf{x}_s = \mathbf{x}_{s+\Delta t} - \Delta t \cdot \mathbf{v}_\theta(\mathbf{x}_{s+\Delta t}, s+\Delta t, p) \text{(w/. grad)},
    \label{eq:gradient_step}
\end{align}
where $\Delta t = \frac{1}{N}$ denotes the discrete step size with $N$ total denoising steps. The process reward loss directly maximizes PAVRM-predicted quality at intermediate timestep $s$:
\begin{equation}
    \mathcal{L}_{\text{PRFL}} = -\lambda \mathbb{E}_{s, p} \left[r_\phi(\mathbf{x}_s, s, p)\right],
    \label{eq:prfl_loss}
\end{equation}
where $\lambda$ controls optimization strength. Crucially, gradients $\nabla_\theta \mathcal{L}_{\text{PRFL}}$ backpropagate through Eq.~(\ref{eq:gradient_step}) and PAVRM $r_\phi$ without requiring VAE decoding $\mathcal{D}(\cdot)$, enabling memory-efficient training.

Following the regularization strategy in Eq.(~\ref{eq:refl_loss}), we prevent reward over-optimization by alternating with supervised fine-tuning on curated dataset $\mathcal{D}_{\text{SFT}} = \{(V_i, p_i)\}_{i=1}^M$:
\begin{equation}
    \mathcal{L}_{\text{SFT}} = \mathbb{E}_{t, (V,p)} \left\|\mathbf{v}_\theta(\mathbf{x}_t, t, p) - (\mathbf{x}_1 - \mathbf{x}_0)\right\|_2^2,
    \label{eq:sft_loss}
\end{equation}
where $\mathbf{x}_0 = \mathcal{E}(V)$, $t \sim \mathcal{U}(0,1)$, and $(V,p) \sim \mathcal{D}_{\text{SFT}}$. This balanced training strategy, alternating between $\mathcal{L}_{\text{PRFL}}$ and $\mathcal{L}_{\text{SFT}}$ at each iteration, maintains generation diversity while adapting to motion quality preferences.

\begin{table*}[htbp]
\centering
\caption{\textbf{Comprehensive comparison across video generation benchmarks on text-to-video generation task.} Evaluation is conducted on Inner Test Set, VBench and VBench2. For Inner Test Set, we evaluate MS (motion smoothness), DD (dynamic degree), SC (subject consistency) from VBench, HA (human anatomy) from VBench2, and our proposed PAVRM. Base model is Wan2.1-T2V-14B. \textbf{Bold} denotes the best results. \textcolor{blue}{Blue} values indicate absolute improvements of PRFL over Pretrain baseline. PRFL achieves significant improvements in dynamic degree and human anatomy.}
\footnotesize
\resizebox{\textwidth}{!}{
\begin{tabular}{llcccccccccccc}
\toprule
\multirow{2}{*}{\textbf{Method}} & \multirow{2}{*}{\textbf{Resolution}} & \multicolumn{5}{c}{\textbf{Inner Test Set}} & \multicolumn{4}{c}{\textbf{VBench}} & \multicolumn{2}{c}{\textbf{VBench2}} & \multirow{2}{*}{\textbf{Avg}} \\
\cmidrule(lr){3-7} \cmidrule(lr){8-11} \cmidrule(lr){12-13}
& & \textbf{MS} & \textbf{DD} & \textbf{SC} & \textbf{HA} & \textbf{PAVRM} & \textbf{MS} & \textbf{DD} & \textbf{SC} & \textbf{PAVRM}&\textbf{HA}&\textbf{PAVRM} & \\
\midrule
Pretrain~\cite{wan2025} & 480P & \textbf{99.20} & 22.00 & \textbf{97.34} & 84.24 & 89.00 & 98.00 & 68.06 & 92.74& 97.22 & 74.38 & 69.17 & 81.03 \\
SFT & 480P & 98.96 & 44.00 & 96.61 & 92.79 & 92.00 & 97.87 & 62.50 & 93.21 & 95.83& 84.80 & 74.17 & 84.79 \\
RWR~\cite{liu2025improving} & 480P & 98.99 & 60.00 & 95.93 & 91.85 & 88.00 & 98.06 & 65.28 & 92.48 & 93.06& 79.67 & 62.50 & 84.17 \\
RGB ReFL~\cite{lin2025contentv} & 480P & \textbf{99.20} & 38.00 & 92.26 & 91.68 & 92.00 & \textbf{98.64} & 62.50 & 90.60 & 97.22& 84.87 & \textbf{76.67} & 83.97 \\
\rowcolor{gray!20}
PRFL & 480P & 99.05 & 68.00 & 96.34 & \textbf{94.73} & 92.00 & 98.18 & 76.39 & 94.16 & \textbf{100.00}& 89.84 & \textbf{76.67} & 89.58 \\
\rowcolor{gray!20}
\textit{vs. Pretrain} & & -0.15 & \textcolor{blue}{+46.00} & -1.00 & \textcolor{blue}{+10.49} & \textcolor{blue}{+3.00} & \textcolor{blue}{+0.18} & \textcolor{blue}{+8.33} & \textcolor{blue}{+1.42} & \textcolor{blue}{+2.78}& \textcolor{blue}{+15.46} & \textcolor{blue}{+7.50} &\textcolor{blue}{+8.55} \\
\midrule
Pretrain~\cite{wan2025} & 720P & 99.09 & 25.00 & 96.69 & 78.73 & 94.00 & 97.70 & 61.11 & 90.63 & 98.61& 68.88 & 62.10 & 79.32 \\
\rowcolor{gray!20}
PRFL & 720P & 98.85 & \textbf{81.00} & 96.09 & 90.89 & \textbf{95.00} & 98.06 & \textbf{84.72} & \textbf{95.46} & \textbf{100.00}& \textbf{90.40} & 66.13 & \textbf{90.60} \\
\rowcolor{gray!20}
\textit{vs. Pretrain} & & -0.24 & \textcolor{blue}{+56.00} & -0.60 & \textcolor{blue}{+12.16} & \textcolor{blue}{+1.00} & \textcolor{blue}{+0.36} & \textcolor{blue}{+23.61} & \textcolor{blue}{+4.83} & \textcolor{blue}{+1.39} &\textcolor{blue}{+21.52} &\textcolor{blue}{+4.03} & \textcolor{blue}{+11.28}\\
\bottomrule
\end{tabular}
}
\label{tab:t2v}
\vspace{-4mm}
\end{table*}

\begin{table*}[htbp]
\centering
\caption{\textbf{Comprehensive comparison across video generation benchmarks on image-to-video generation task.} Evaluation is conducted on Inner Test Set and VBench. For Inner Test Set, we evaluate MS (motion smoothness), DD (dynamic degree), SC (subject consistency), IC (i2v subject) from VBench-I2V, and our proposed PAVRM. \textbf{Bold} denotes the best results. \textcolor{blue}{Blue} values indicate absolute improvements of PRFL over Pretrain baseline. PRFL achieves significant improvements in dynamic degree.}

\footnotesize
\resizebox{\textwidth}{!}{
\begin{tabular}{llcccccccccccc}
\toprule
\multirow{2}{*}{\textbf{Method}} & \multirow{2}{*}{\textbf{Backbone}} & \multicolumn{5}{c}{\textbf{Inner Test Set}} & \multicolumn{5}{c}{\textbf{VBench-I2V}} & \multirow{2}{*}{\textbf{Avg}} \\
\cmidrule(lr){3-7} \cmidrule(lr){8-12}
& & \textbf{MS} & \textbf{DD} & \textbf{SC} & \textbf{IC} & \textbf{PAVRM} & \textbf{MS} & \textbf{DD} & \textbf{SC} &\textbf{IC}&\textbf{PAVRM} &\\
\midrule
Pretrain~\cite{wan2025} & Wan2.1-I2V-14B-480P & 98.66 & 57.00 & 91.73 & 96.86 & 87.00 & 97.86 & 40.65 & 93.86 & 97.21& 92.28& 85.31 \\
\rowcolor{gray!20}
PRFL & Wan2.1-I2V-14B-480P & 98.88 & \textbf{87.00} & \textbf{93.18} & 97.31 & \textbf{93.00} & 98.04 & \textbf{81.30} & 94.57 & 97.79 & 92.68& \textbf{93.38} \\
\rowcolor{gray!20}
\textit{vs. Pretrain} & & \textcolor{blue}{+0.22} & \textcolor{blue}{+30.00} & \textcolor{blue}{+1.45} & \textcolor{blue}{+0.45} & \textcolor{blue}{+6.00} & \textcolor{blue}{+0.18} & \textcolor{blue}{+40.65} & \textcolor{blue}{+0.71} & \textcolor{blue}{+0.58} & \textcolor{blue}{+0.40}&\textcolor{blue}{+8.07} \\
\midrule
Pretrain~\cite{wan2025} & Wan2.1-I2V-14B-720P & 98.40 & 60.00 & 90.96 & 96.65 & 74.00 & 98.04 & 35.37 & 94.49 & 97.92 & 89.43& 83.53 \\
\rowcolor{gray!20}
PRFL & Wan2.1-I2V-14B-720P & \textbf{99.03} & 76.00 & 92.83 & \textbf{98.26} & 90.00 & \textbf{98.65} & 68.42 & \textbf{95.62} & \textbf{98.73}& \textbf{95.53}& 91.31 \\
\rowcolor{gray!20}
\textit{vs. Pretrain} & & \textcolor{blue}{+0.63} & \textcolor{blue}{+16.00} & \textcolor{blue}{+1.87} & \textcolor{blue}{+1.61} & \textcolor{blue}{+16.00} & \textcolor{blue}{+0.61} & \textcolor{blue}{+33.05} & \textcolor{blue}{+1.13} & \textcolor{blue}{+0.81} & \textcolor{blue}{+6.10} &  \textcolor{blue}{+7.78} \\
\bottomrule
\end{tabular}
}
\label{tab:i2v}
\vspace{-4mm}
\end{table*}

\begin{figure}
    \centering
    \includegraphics[width=0.9\linewidth]{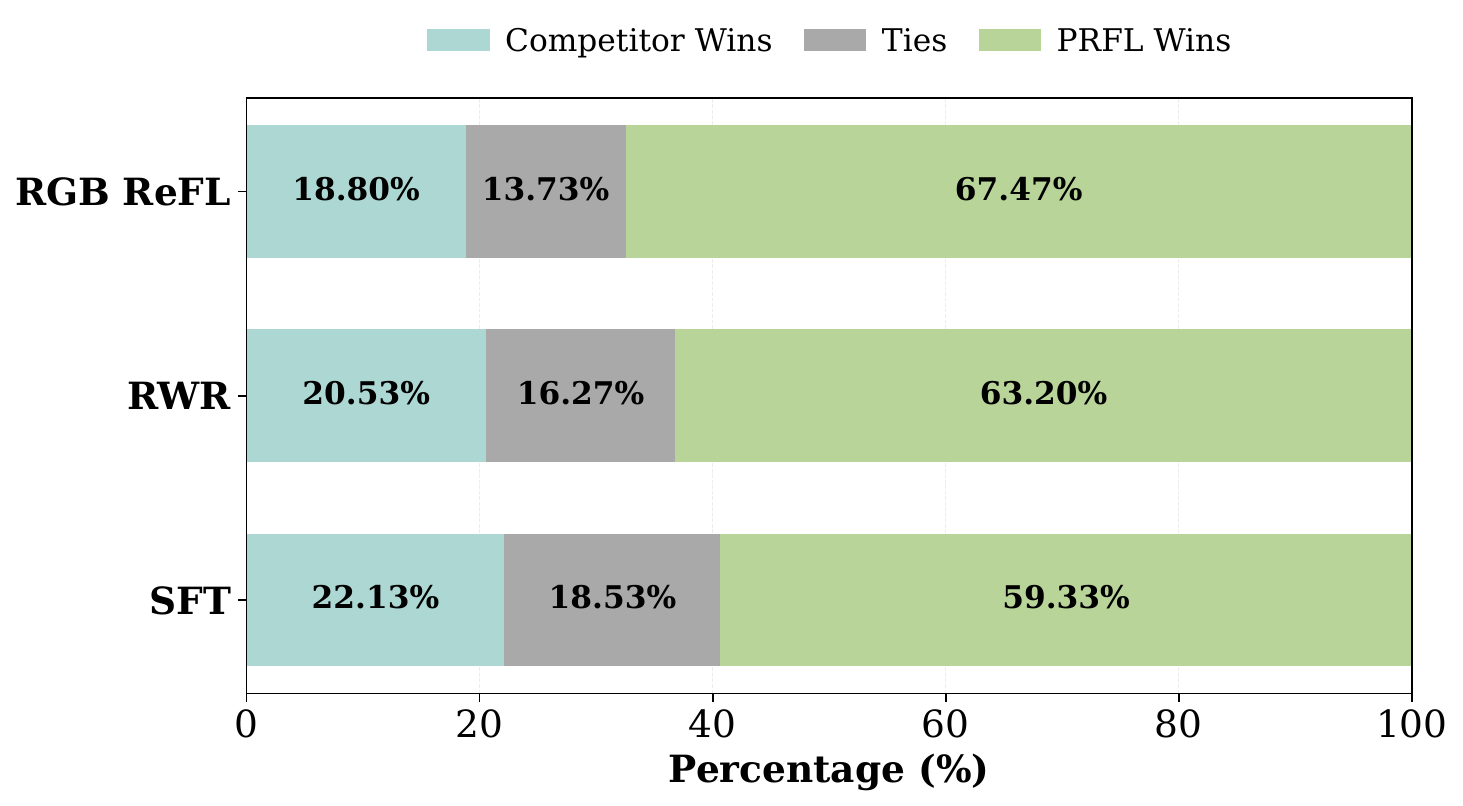}
    \vspace{-2mm}
    \caption{Human evaluation of PRFL
model vs. other post-training methods.}
    \label{fig:user_study}
    \vspace{-4mm}
\end{figure}
\begin{figure*}
    \centering
    \includegraphics[width=1\linewidth]{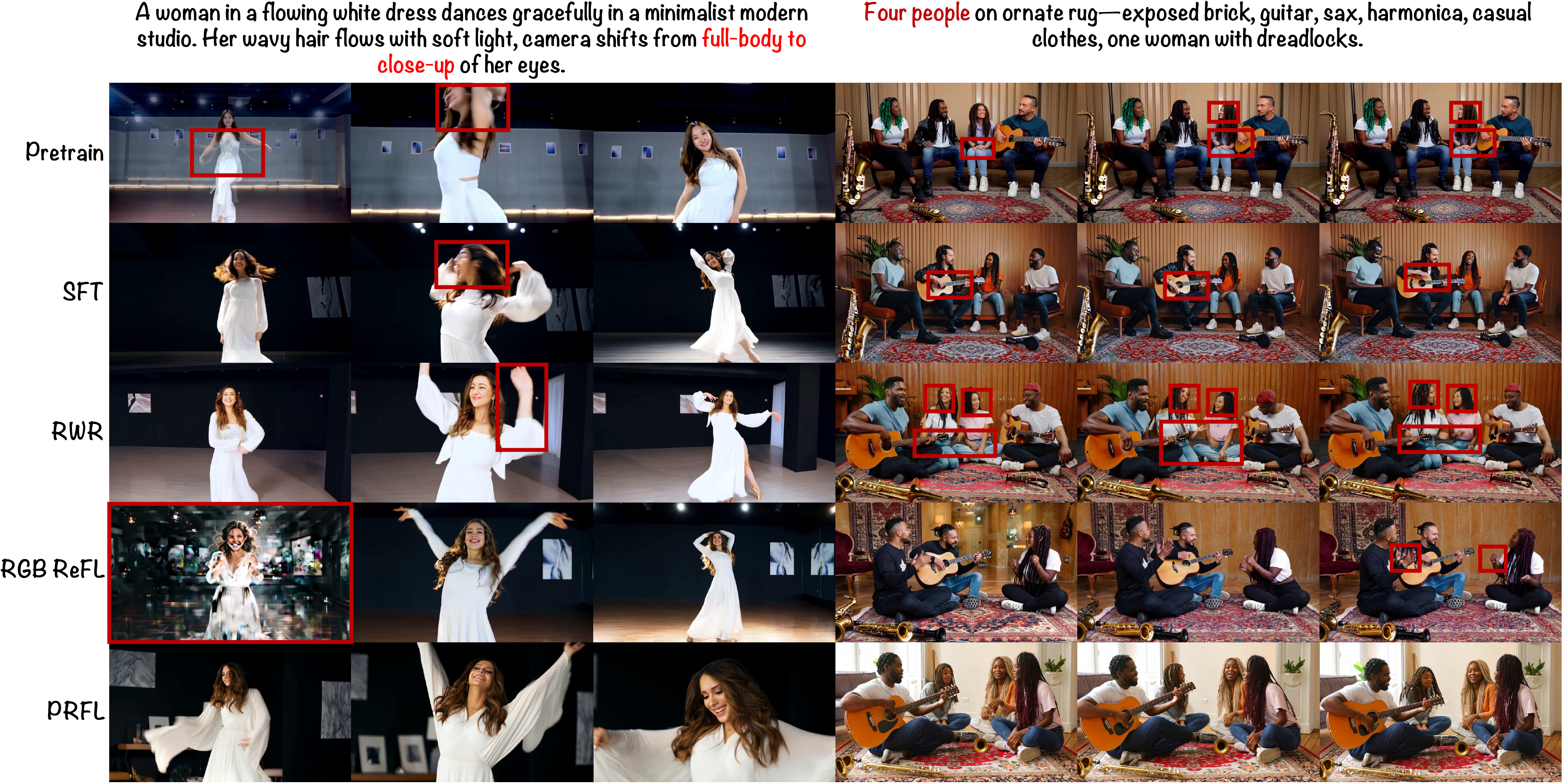}
    \caption{Qualitative results for different post-training methods on 480P text-to-video task. The \textcolor{red}{red box} highlights the generated artifacts.
    Zoom in for a better view.
    For the complete prompt, please see the supplementary materials.}
    \label{fig:case_1}
\end{figure*}
\begin{table*}[htbp]
\centering
\vspace{-4mm}
\caption{\textbf{Analysis of the influence of different timestep-sampling strategies.}
Base model is Wan2.1-T2V-14B at 480P resolution. The early, middle, and late stages of the denoising process refer to the first third, middle third, and final third of the process, respectively.
}
\footnotesize
\begin{tabular}{lcccccccccccc}
\toprule
\multirow{2}{*}{\textbf{Method}} & \multicolumn{5}{c}{\textbf{Inner Test Set}} & \multicolumn{4}{c}{\textbf{VBench}} & \multicolumn{2}{c}{\textbf{VBench2}} & \multirow{2}{*}{\textbf{Avg}} \\
\cmidrule(lr){2-6} \cmidrule(lr){7-10} \cmidrule(lr){11-12}
& \textbf{MS} & \textbf{DD} & \textbf{SC} & \textbf{HA} & \textbf{PAVRM} & \textbf{MS} & \textbf{DD} & \textbf{SC} & \textbf{PAVRM}&\textbf{HA}&\textbf{PAVRM} & \\
\midrule
Pretrain & \textbf{99.20} & 22.00 & \textbf{97.34} & 84.24 & 89.00 & 98.00 & 68.06 & 92.74& 97.22 & 74.38 & 69.17 & 81.03 \\
Early Stage &  99.00 & 51.00 & 96.63 & 87.52 & 85.00 & 98.53 & 48.61 &\textbf{ 96.71} & 95.83&78.47 & 67.50&82.25 \\
Middle Stage & 99.11 & 51.00 & 96.69 & 89.38 & 92.00 & 98.32 &\textbf{ 76.39} & 94.84 & 98.61&80.92 &80.00& 87.02 \\
Late Stage  & \textbf{99.26} & 44.00 & 96.29 & 91.54 & \textbf{93.00} & \textbf{98.60} & 58.33 & 93.94 &98.61& 84.07 & 77.50&85.01 \\
Full Stage  & 99.05 & \textbf{68.00} & 96.34 & \textbf{94.73} & 92.00 & 98.18 & \textbf{76.39} & 94.16 & \textbf{100.00}& 89.84 & \textbf{76.67} & \textbf{89.58} \\

\bottomrule
\end{tabular}
\label{tab:analysis_timestep}
\vspace{-4mm}
\end{table*}

\begin{table}[htbp]
  \centering
  \caption{Comparison of training resource consumption and efficiency. PRFL achieves 1.4× speedup while processing full frames. Here \textsuperscript{\textdagger} means without SFT loss, and \textbf{bold} means the best performance.}
  \vspace{-4mm}
  \label{tab:training_resource_efficiency}
  \footnotesize
  \setlength{\tabcolsep}{.5mm}
  \begin{tabular}{lccc}
    \toprule
    \textbf{Method} & \textbf{VRAM (GB)} & \textbf{Time per Step (s)} & \textbf{Speedup} \\
    \midrule
    RGB ReFL (full frames)  & OOM & - & - \\
    RGB ReFL (first frame) &\textbf{ 55.47} & 72.38 / 64.89\textsuperscript{\textdagger} & 1.00$\times$ \\
    \rowcolor{gray!20}
    PRFL (full frames) & 66.81 & \textbf{51.11 / 43.69\textsuperscript{\textdagger}} & \textbf{1.42$\times$ / 1.49$\times$} \\
    \bottomrule
  \end{tabular}
  \vspace{-6mm}
\end{table}
\begin{table*}[htbp]
\centering
\caption{Quantitative results for different architecture PAVRMs across different timesteps on T2V and I2V tasks under 480P and 720P. The metric is average accuracy calculated after randomly sampling a time step within each time step range for every test sample. \textbf{Bold} means best performance.}
\label{tab:quan_LRMs}
\footnotesize
\begin{tabular}{lcccccccc}
\toprule
\textbf{Method} &\textbf{Base Model}& \textbf{Task} & \textbf{[0, 0.2]} & \textbf{(0.2, 0.4]} & \textbf{(0.4, 0.6]} & \textbf{(0.6, 0.8]} & \textbf{(0.8, 1.0]} & \textbf{Avg} \\
\midrule
\multicolumn{8}{l}{\textit{Resolution: 720P}} \\
VideoAlign~\cite{liu2025improving}&VLM&I2V &- & - & - & - & - & 78.83 \\
VideoPhy~\cite{bansal2025videophy}&VLM&I2V &- & - & - & - & - & 77.04 \\
Mean Pooling & VGM&I2V & 82.40 & 82.91 & 83.67 & \textbf{85.46} & 82.40 & 83.37 \\
Max Pooling & VGM&I2V & 80.61 & 80.10 & 80.61 & 77.30 & 73.72 & 78.47 \\
Attention w./o. query &VGM& I2V & 80.36 & 84.44 & \textbf{84.95} & 84.18 & 83.16 & 83.42 \\
\rowcolor{gray!20} Attention w./ query &VGM& I2V & 83.42 & \textbf{84.95} & 84.69 & 84.44 & 83.42 & \textbf{84.18} \\
\rowcolor{gray!20} Attention w./ query &VGM& T2V & 82.40 & 84.69 & 84.69 & 84.44 & \textbf{84.44} & 84.13 \\
\midrule
\multicolumn{8}{l}{\textit{Resolution: 480P}} \\
\rowcolor{gray!20} Attention w./ query & VGM&I2V & \textbf{83.67} & 84.18 & 84.69 & 84.69 & 79.85 & 83.42 \\
\midrule
\rowcolor{gray!20} Attention w./ query & VGM&T2V & 81.89 & 83.93 & 84.18 & 83.93 & 83.16 & 83.42 \\
\bottomrule
\end{tabular}
\end{table*}

\section{Experiments}
\label{sec:exp}
\subsection{Experimental Setup}
\paragraph{Dataset.}
We collected about 31,000 portrait videos from online sources and generated text prompts using a video caption model. The first frame and text prompt of each video are fed into the Wan2.1-14B-I2V model for inference. The generated videos are labeled by professional annotators based on motion quality as qualified, partially qualified, and unqualified. The partially qualified videos are filtered out to enhance distinctiveness. Detailed annotation and filtration guidelines are provided in the supplementary materials. This process yielded a dataset comprising 24,000 real videos and corresponding generated videos.
For reward model, we utilized only the generated videos, and randomly selected 500 samples as the test set (100 for validation, 400 for testing). For video generation, we employed only real videos as SFT training data, and we randomly selected 100 input conditions as our test dataset from the 500-sample test set of the reward model.
We also incorporated the existing open-source benchmark VBench~\cite{huang2023vbench} and VBench2~\cite{zheng2025vbench2} to ensure a fair comparison with state-of-the-art methods.

\vspace{-4mm}
\paragraph{Training Details.}
We adopt Wan2.1-I2V-14B~\cite{wan2025} as our primary baseline. We employ the AdamW optimizer~\cite{loshchilov2017decoupled} with learning rates of 1e-5 for query attention and head of PAVRM, 1e-6 for the feature extraction of PAVRM, and 5e-6 for PRFL. We utilize UniPCMultistepScheduler~\cite{zhao2023unipc} with 1,000 training steps and 40 inference steps.

\vspace{-4mm}
\paragraph{Comparison.}
For reward model, we select VideoAlign~\cite{liu2025improving} and VideoPhy~\cite{bansal2025videophy}, a state-of-the-art RGB-based reward model. 
For video generation, we compare against four post-training methods, two offline methods: SFT~\cite{liu2023flow} and reward weighted regression~(RWR)~\cite{liu2025improving}, and one online RL method RGB ReFL~\cite{lin2025contentv}, which only decodes the first frame and uses PickScore~\cite{kirstain2023pick} as reward model.
See details in supplementary materials.
\vspace{-4mm}
\paragraph{Evaluation.}
For PAVRM evaluation, we conduct stratified random sampling of timestep t across five intervals: [0, 0.2], (0.2, 0.4], (0.4, 0.6], (0.6, 0.8], and (0.8, 1.0]. Each test sample is randomly sampled once within each interval, and the final average accuracy is computed. 
For video generation evaluation, generated videos are produced at 720P/480P resolution.
For the text-to-video generation task, we employ evaluation metrics from the VBench series, including dynamic degree, motion smoothness, subject consistency, and Human Anatomy, as well as the qualified ratio across the whole test set estimated by PAVRM. Specifically, for each test sample, we random sample a timestep t across [0, 1.0] fed into PAVRM and get predictions 0 as unqualified or 1 as qualified. 
For the image-to-video generation task, additional i2v subject metric is used from the VBench i2v series.

\subsection{Video Generation}
\paragraph{Quantitative Results.}
As shown in Tables~\ref{tab:t2v} and~\ref{tab:i2v}, we evaluate PRFL across different resolutions and tasks.

\noindent\textit{Text-to-Video Generation.}
In the T2V 480P setting, PRFL substantially outperforms other post-training methods (SFT, RWR, RGB ReFL) in dynamic degree (+46.00 on Inner Test Set) and human anatomy (+10.49), while maintaining high performance in motion smoothness (99.05) and subject consistency (96.34). The slight decreases in MS (-0.15) and SC (-1.00) are marginal given their already high baseline values (99.20 and 97.34). This suggests PRFL enhances motion dynamics without significantly compromising smoothness or consistency. Similar trends are observed at 720P resolution, with even larger improvements in dynamic degree (+56.00) and human anatomy (+12.16).

\noindent\textit{Image-to-Video Generation.}
PRFL generalizes well to the I2V task. At 480P, dynamic degree improves by +30.00 on Inner Test Set and +40.65 on VBench-I2V, while other metrics either improve or remain stable (MS: +0.22, SC: +1.45, IC: +0.45). The 720P results confirm consistent improvements across metrics. These results indicate that PRFL's effectiveness extends beyond T2V generation.

The consistent improvements in motion-related metrics (dynamic degree, human anatomy) across different settings, combined with the preservation of high-level smoothness and consistency, demonstrate the effectiveness of PRFL. While automatic metrics have limitations in capturing perceptual quality, the quantitative results provide evidence that PRFL successfully optimizes motion quality across diverse video generation scenarios.
\vspace{-4mm}
\paragraph{User Study.}
To complement automatic metrics, we conduct a comprehensive human evaluation study. We randomly sample 25 prompts from the test set and generate videos using PRFL and three baseline methods (SFT, RWR, RGB ReFL). We recruit 2,250 pairwise comparisons based on overall video quality from 30 professional participants.
Each participant evaluates video pairs and chooses A wins/tie/B wins based on comprehensive consideration of text instruction adherence, visual quality, and motion quality.
As shown in Fig.~\ref{fig:user_study}, PRFL consistently outperforms all baseline methods. 
These results demonstrate that PRFL's process-level optimization effectively improves video quality as perceived by human evaluators. 
\vspace{-4mm}
\paragraph{Qualitative Results.}
As shown in Fig.~\ref{fig:case_1}, we visualize three representative frames (early, middle, and late stages) from generated videos to compare PRFL with baseline methods.
In the dancing scenario with complex motion dynamics, baseline methods exhibit various artifacts: Pretrain generates a distorted environment and figure, SFT produces facial distortions in the close-up frame, RWR shows body deformations, and RGB ReFL generates a failed first frame.
In contrast, PRFL generates consistent and artifact-free frames throughout the video while maintaining smooth motion transitions and accurately following the text prompt's camera movement requirements. In the multi-person guitar scene, baseline methods struggle with anatomical correctness (e.g., distorted hands and faces highlighted in red boxes), whereas PRFL maintains visual quality and generates anatomically plausible human figures and objects across all frames. These qualitative comparisons demonstrate that PRFL effectively reduces generation artifacts while preserving motion coherence in challenging scenarios.
We also provide many human and non-human video cases in the supplementary materials.
\vspace{-4mm}
\paragraph{Sensitivity to Sampling Timesteps.}
We analyze how different denoising stages affect generation quality by training PRFL with timestep sampling restricted to early, middle, and late stages of the denoising process, comparing against full-range sampling. As shown in Table~\ref{tab:analysis_timestep}, early and middle stages primarily govern dynamic degree and motion quality—both achieve DD of 51.00 on Inner Test Set, with the middle stage showing stronger VBench improvements (76.39 vs. 48.61). Late-stage sampling contributes less to dynamics (DD: 44.00) but substantially improves human anatomy (HA: 84.24→91.54). Full-range sampling achieves optimal overall performance (Avg: 89.58, DD: 68.00, HA: 94.73), demonstrating that optimizing across all denoising stages is essential for balancing motion and structure quality.
\vspace{-4mm}
\paragraph{Computational Efficiency.}
As shown in Table~\ref{tab:training_resource_efficiency}, we measure the peak VRAM consumption and average time per training step for different ReFL algorithms, both with and without the SFT loss component.
The results demonstrate that PRFL can decode all 81 frames while maintaining peak VRAM usage within 67 GB, which is practical for most mainstream GPU clusters. Notably, RGB-ReFL encounters out-of-memory errors when attempting to process full 81-frame sequences and must resort to first-frame-only training. Despite processing significantly more visual information (81 frames vs. 1 frame), PRFL achieves a 1.42$\times$ to 1.49$\times$ training speedup compared to the first-frame-only RGB-ReFL baseline.

\subsection{Reward Model}
\paragraph{Architecture Selection and Timestep Analysis.}
As shown in Table~\ref{tab:quan_LRMs}, we evaluate the accuracy of PAVRM across different denoising stages and compare various aggregation methods. Attention w./ query achieves the best average performance (84.18\% for I2V at 720P) and maintains stable accuracy across all timestep ranges, while simpler methods like Max Pooling suffer significant degradation in later stages. Our reward model training approach demonstrates strong generalization across different tasks (T2V vs. I2V), resolutions (480P vs. 720P), and timestep ranges, maintaining over 83\% average accuracy in all evaluated settings.
\section{Conclusion}
\label{sec:conclusion}
We present an RLHF framework for aligning video generation models with human preferences through process-level reward modeling and optimization. Our Process-Aware Video Reward Model (PAVRM) evaluates motion quality directly from noisy latent representations at arbitrary timesteps using query-based aggregation for variable-length videos. Process Reward Feedback Learning (PRFL) enables memory-efficient fine-tuning by operating in latent space, eliminating VAE decoding overhead. Experiments demonstrate substantial motion quality improvements with large memory reduction and at least 1.4$\times$ training speedup versus pixel-space methods.
\vspace{-4mm}
\paragraph{Limitations and Future Work.}
Our method focuses on motion quality and may benefit from multi-aspect evaluation covering aesthetics and semantics. Future work could explore hybrid reward frameworks, richer preference signals, and extensions to controllable generation and video editing.
{
    \small
    \bibliographystyle{ieeenat_fullname}
    \bibliography{main}
}

\clearpage
\appendix
\setcounter{page}{1}

\twocolumn[{
\maketitle
\begin{center}
{\LARGE \textbf{Appendix}}\\
\vspace{0.5cm}
{\Large Video Generation Models are Good Latent Reward Models}
\vspace{1cm}
\end{center}
}]
\renewcommand{\thesection}{A.\arabic{section}}
\setcounter{section}{0}
\renewcommand{\thesubsection}{\thesection.\arabic{subsection}}
\numberwithin{equation}{section}
\section{Overview.}
We provide detailed experimental settings in~\cref{sec:detail_ex_setting}, and more experimental analysis about our reward model in~\cref{sec:exp_more}.
\section{More Details in Experimental Settings}
\label{sec:detail_ex_setting}
\subsection{Analysis Experiments}
We selected the I2V task and the resolution of 720P for our analysis experiments, with the same dataset of our PAVRM training.
The linear probe layer dimension matches the token dimension of the VGM (5120 for Wan2.1-I2V-14B).

For Fig.~\ref{fig:analysis}(a), we performed standard 40-step inference on dataset samples, denoise each intermediate timestep noisy latent to clean latent by one step, and decoding clean latent of each intermediate timestep to RGB space for storage. We computed the average score of dataset samples of each timestep using VideoAlign-MQ, a state-of-the-art VLM-based Video Reward Model. The results reveal substantial divergence between high-noise regions and clean videos, demonstrating that RGB-based video reward models fail to directly generalize as latent reward models.

For Fig.~\ref{fig:analysis}(b), we fixed the timestep at $t=0.2$ and analyzed the impact of varying DiT block counts on VGM performance as a reward model. To isolate the influence of VGM features on video quality assessment, we employed mean pooling for feature aggregation. The linear probe was trained identically to our PAVRM using BCE loss. For accuracy calculation, we applied a threshold of 0 to the linear probe output: predictions $p_r\geq 0$ were classified as good videos, otherwise as bad videos. The VideoAlign test accuracy of 78.83\% was obtained by setting a threshold on VideoAlign MQ reward scores—videos with scores above the threshold were labeled as good, otherwise bad—and computing accuracy against ground-truth labels. This threshold was selected to maximize test set accuracy.

For Fig.~\ref{fig:analysis}(c), ``Fixed $t$ (MLP-only)'' refers to training only the linear probe at fixed timesteps ($t=0.2/0.4/0.6/0.8$), resulting in four separate models with test accuracy computed using the same way as (b), based on 8 DiT blocks. ``Random $t$ (MLP-only)'' involves training a single model where timesteps are randomly sampled during training using UniPCMultistepScheduler with 1000 training steps. ``Random $t$ (Full fine-tuning)'' fine-tunes both DiT blocks and the linear probe (excluding text/image encoders and VAE), with all other settings identical to ``Random $t$ (MLP-only)''. 

\subsection{Open Source Test Set}
We incorporated the existing open-source benchmark VBench~\cite{huang2023vbench} and VBench2~\cite{zheng2025vbench2} to ensure a fair comparison with state-of-the-art methods.
In this paper, we validate the effectiveness of our method in terms of motion quality, which requires the video generation process to be free of distortions, exhibit smooth motion, and comply with physical laws.
For the text-to-video (T2V) task, we selected the subject consistency subset from VBench, a total of 72 prompts.
Additionally, we employed the human anatomy subset from VBench2, which includes the human anatomy metric with 120 prompts specifically enhanced for the Wan model.
For the image-to-video (I2V) task, we selected I2V Subject subset from VBench-I2V (in VBench++~\cite{huang2024vbench++}), a total of 246 prompts.
\subsection{Inner Data Collection and Annotations}
\paragraph{Data Generation Pipeline.}
Our inner dataset construction begins with an internal collection of 31000 high-quality human portrait videos. Using the first frame and corresponding text prompt from each video, we generated synthetic videos using the Wan2.1-14B-I2V model. Due to computational constraints, we generated one 720P video per input condition, with each video requiring approximately 30 minutes of inference time on a single GPU.
\paragraph{Annotation Protocol.}
The annotation process consists of two stages: automatic filtering and manual quality assessment.
\noindent\textit{Stage 1: Coarse Filtering.}
We first removed videos exhibiting obvious defects, including black screens, or visible watermarks.
\begin{table*}[!ht]
\centering
\caption{Definitions of evaluation dimensions and assessment criteria used in our human annotation framework.}
\label{tab:evaluation_dimensions_styled}
\begin{tabular}{c|c}
\toprule
\textbf{Evaluation Dimension} & \textbf{Definition and Assessment Criteria}  \\
\midrule
\textbf{Physical Plausibility} & \begin{tabular}{p{0.7\textwidth}}
Evaluates whether video dynamics adhere to \textbf{real-world physical principles}.
\begin{itemize}[label={-}, leftmargin=15pt, topsep=3pt, partopsep=0pt, itemsep=2pt]
        \item \textbf{Motion Dynamics:} Assesses whether object motion exhibits realistic acceleration, deceleration, and inertia consistent with natural physics.
        \item \textbf{Interaction Realism:} Evaluates the plausibility of physical interactions, including gravitational effects (e.g., falling objects), collision dynamics, and force propagation (e.g., splashing water).
        \item \textbf{Material Behavior:} Examines the realistic deformation and dynamics of complex materials, including fluid motion (water, smoke) and soft body dynamics (cloth, skin).
    \end{itemize}
    \end{tabular} \\
    \midrule
    \textbf{Subject Deformity} & \begin{tabular}{p{0.7\textwidth}}
    Assesses \textbf{structural integrity and temporal consistency} of subjects (humans, animals, objects).
    
    \begin{itemize}[label={-}, leftmargin=15pt, topsep=3pt, partopsep=0pt, itemsep=2pt]
        \item \textbf{Structural Integrity:} Evaluates anatomical correctness and structural coherence, penalizing severe distortions, unnatural proportions, or implausible body parts (e.g., malformed faces, extra limbs).
        \item \textbf{Temporal Consistency:} Measures the stability of subject identity and form across frames, penalizing artifacts such as shape morphing, flickering, melting effects, or sudden appearance changes.
    \end{itemize}
    \end{tabular} \\
    \bottomrule
\end{tabular}
\end{table*}
\noindent\textit{Stage 2: Manual Quality Assessment.}
The remaining videos were manually annotated by professional annotators across two key dimensions: \textbf{Physical Plausibility} and \textbf{Subject Deformity}. Each dimension was rated using a three-level scale: qualified, partially qualified, and unqualified. The specific criteria for each rating level are detailed in Table~\ref{tab:evaluation_dimensions_styled}.
For Physical Plausibility:
\begin{itemize}[leftmargin=15pt, topsep=3pt, itemsep=2pt]
\item \textit{Qualified}: Motion appears smooth and natural, following real-world physics with realistic acceleration, deceleration, and interactions.
\item \textit{Partially Qualified}: Minor physical inconsistencies exist but do not severely impact overall believability.
\item \textit{Unqualified}: Significant violations of physical laws, such as objects defying gravity, unnatural motion trajectories, or implausible interactions.
\end{itemize}
For Subject Deformity:
\begin{itemize}[leftmargin=15pt, topsep=3pt, itemsep=2pt]
\item \textit{Qualified}: Subjects maintain consistent structure and identity throughout the video with no visible artifacts.
\item \textit{Partially Qualified}: Minor temporal inconsistencies or subtle structural artifacts that do not fundamentally distort the subject.
\item \textit{Unqualified}: Severe anatomical distortions, identity shifts, or temporal artifacts such as melting, flickering, or morphing.
\end{itemize}
\paragraph{Label Construction.}
To enhance data distinctiveness and reduce annotation ambiguity, we applied the following labeling strategy: videos rated as \textit{qualified} on both dimensions were labeled as \textbf{good videos}, while those rated as \textit{unqualified} on both dimensions were labeled as \textbf{bad videos}. Videos with mixed ratings (e.g., qualified on one dimension but unqualified on another) or those marked as \textit{partially qualified} on either dimension were excluded from the final dataset to ensure clear decision boundaries.
\paragraph{Validation and Test Set Construction.}
We randomly sampled 500 videos for evaluation purposes: 100 for validation and 400 for testing. To ensure annotation reliability, each sample in both the validation and test sets was independently annotated by at least three professional annotators, with final labels determined by majority voting.
\paragraph{Final Dataset Statistics.}
After filtering and annotation, our final dataset comprises 24000 video pairs (real and generated), with the generated videos used for reward model training and the real videos used as supervised fine-tuning (SFT) data for video generation. The dataset distribution is as follows: approximately 23500 samples for training, 100 for validation, and 400 for testing the reward model.

\subsection{Baseline Settings}
For reward models, we select VideoAlign-MQ~\cite{liu2025improving} and VideoPhy-PC~\cite{bansal2025videophy}, two state-of-the-art VLM-based reward models that excel particularly in assessing motion quality. Both models are employed in a zero-shot manner. The accuracy (Acc) metric is computed by establishing a threshold on the reward scores: videos with scores at or above the threshold are classified as ``good'', while those below are classified as ``bad''. The accuracy is then calculated against ground-truth labels, where the reported threshold is selected to maximize accuracy on the test set.

For post-training, we utilize approximately the same number of training samples across all methods, performing one epoch over the text-video pairs with a sequence parallel size of 4, with same learning rate of 5e-6 and a global batch size of 30 (i.e. batch size of 6 with gradient accumulation number of 5).

\paragraph{Supervised Fine-Tuning (SFT).}
SFT~\cite{liu2023flow} is a widely adopted and effective post-training technique that offers high computational efficiency. From a reinforcement learning perspective, it can be viewed as an offline, off-policy algorithm, optimizing the loss function defined in Equation~\ref{eq:flow_loss}.

\paragraph{Reward Weighted Regression (RWR).}
Reward weighted regression (RWR)~\cite{liu2025improving} is a prevalent and effective offline, off-policy RL method that has demonstrated success across traditional RL tasks~\cite{peng2019advantage}, image generation~\cite{furuta2024improving}, and video generation~\cite{liu2025improving}.
RWR directly learns from pre-sampled training data treated as experience samples, where a reward model scores each sample to determine its weight in the training loss.
The loss function is given by:
\begin{equation}
    \begin{split}
        \mathcal{L}_{\text{RWR}}(\theta) &= \mathbb{E}_{t \sim \mathcal{U}(0,1),\;\mathbf{x}_0\sim q(\mathbf{x}_0),\;\mathbf{x}_1\sim p(\mathbf{x}_1)} \bigl[ \\
        &\quad \exp({r_{\phi}(\text{video},y)})\|\mathbf{v}_\theta(\mathbf{x}_t, t) - \mathbf{v}\|^2\bigr],
    \end{split}
    \label{eq:rwr_loss}
\end{equation}
Following the VideoAlign framework, we utilize VideoAlign-MQ to provide reward signals with varying weight configurations.

\paragraph{RGB ReFL.}
For RGB ReFL, we adopt the implementation from ContentV~\cite{lin2025contentv}, which performs VAE decoding only on the first frame and employs the image reward model PickScore~\cite{kirstain2023pick}. The loss function is as follows:
\begin{equation}
\label{eq:rgb_refl_loss}
\mathcal{L}_{\text{RGB ReFL}} = -\lambda \mathbb{E}_{\mathbf{x}_0 \sim \text{VGM}_\theta} \left[ r_\phi(\mathcal{D}(\mathbf{x}_0')) \right] + \mathcal{L}_{\text{FM}}(\theta),
\end{equation}
where $\mathbf{x}_0'$ denotes the latent feature corresponding to the first frame.

\subsection{Evaluation}

\paragraph{Experimental Configuration.}
Following the standardized protocol recommended by Wan2.1, we generate evaluation videos at 720P/480P resolution. During the inference phase of video generation models, we maintain a classifier-free guidance (CFG) weight of 5.5. The sampling process employs the UniPCMultistepScheduler~\cite{zhao2023unipc} over 40 iterative steps. The early, middle, and late stages of the denoising process correspond to steps 1-13, 14-26, and 27-40, respectively.

\paragraph{Automatic Evaluation Metrics.}
To assess the performance of our reward model, we implement a stratified sampling approach across the temporal dimension. The timestep $t$ is partitioned into five uniform intervals: [0, 0.2], (0.2, 0.4], (0.4, 0.6], (0.6, 0.8], and (0.8, 1.0]. Within each interval, test samples undergo random sampling exactly once, and the reward accuracy metric is derived by averaging the accuracy across all intervals.

For text-to-video generation tasks, we adopt multiple evaluation dimensions inspired by the VBench framework, encompassing dynamic degree, motion smoothness, subject consistency, and human anatomy accuracy. In image-to-video generation scenarios, we additionally incorporate the image-video subject consistency metric. Furthermore, we utilize the PAVRM score to quantify the proportion of qualified samples across the entire test set.

\noindent\textit{Motion Smoothness.} Following VBench, we evaluate motion fluidity using frame interpolation priors. Given a video sequence $[f_0, f_1, \ldots, f_{2n}]$, we remove odd-indexed frames to create $[f_0, f_2, \ldots, f_{2n}]$, then reconstruct the missing frames $[\hat{f}_1, \hat{f}_3, \ldots, \hat{f}_{2n-1}]$ via interpolation. The normalized MAE between reconstructed and original frames yields a score in $[0, 1]$, with higher values indicating smoother motion.

\noindent\textit{Dynamic Degree.} To measure generation dynamism, we adopt VBench's approach using RAFT~\cite{teed2020raft} to estimate inter-frame optical flow. We compute the mean of the top 5\% flow magnitudes as a static/dynamic threshold, with the final score representing the proportion of non-static videos generated.

\noindent\textit{Subject Consistency.} We adopt VBench's DINO-based~\cite{caron2021emerging} metric to assess subject identity preservation across frames. The consistency score is:
\begin{equation}
    S_{\text{subject}} = \frac{1}{T-1}\sum_{t=2}^{T} \frac{1}{2}\left(\langle d_1, d_t \rangle + \langle d_{t-1}, d_t\rangle\right),
\end{equation}
where $d_i$ is the normalized DINO feature of frame $i$, and $\langle \cdot, \cdot \rangle$ computes cosine similarity. This jointly measures consistency with the first frame and temporal continuity.

\noindent\textit{Human Anatomy.} We use the VBench metric that they train three ViT-based~\cite{xie2022simmim} anomaly detectors for human torso, hands, and faces. Training data includes $\sim$1K real videos (YOLO-World~\cite{cheng2024yolo} extracted patches as positives) and $\sim$1K synthetic videos from CogVideo~\cite{hong2022cogvideo,yang2024cogvideox} and HunyuanVideo~\cite{kong2024hunyuanvideo}, plus HumanRefiner~\cite{fang2024humanrefiner} negatives, totaling $\sim$150K annotated frames. The score is the percentage of frames without detected anomalies.

\noindent\textit{I2V Subject Consistency.} We use the VBench++~\cite{huang2024vbench++} metric to evaluate input image-to-video subject correspondence. DINOv1~\cite{caron2021emerging} features are extracted from the input image and video frames. The final score combines weighted similarities between the input image and each frame, plus inter-frame similarities, addressing variations in how models handle input images.

\noindent\textit{PAVRM Score.} To estimate the qualified sample ratio across the test set, we adopt a randomized evaluation protocol. For each test sample, we randomly sample a timestep $t$ from the interval $[0, 1.0]$ and feed it to the PAVRM model, which produces a binary prediction: 0 for unqualified and 1 for qualified. The overall qualified ratio serves as the PAVRM score metric.
\paragraph{Complete Prompt in Fig.\ref{fig:case_1}}
\noindent \textit{Case 1.}
A woman in a flowing white dress is dancing gracefully in a modern dance studio. Her movements are fluid and expressive, with arms sweeping widely and legs moving in elegant, rhythmic patterns. She has long wavy hair that flows freely with each movement, catching the soft lighting from above. The background is a minimalist setup with black walls and a few abstract paintings hanging on them. The camera follows her from a medium shot, capturing her full body as she dances, then moves to a close-up of her face, highlighting her joyful expression and the sparkle in her eyes. The video has smooth transitions and dynamic camera movements, including tracking shots and slow-motion sequences to emphasize her graceful movements.

\noindent \textit{Case 2.}
Four people are seated on an ornate rug in a room with exposed brick walls. One man holds an acoustic guitar. Instruments including a saxophone and harmonica rest on the floor near them. The individuals have varying hair colors and styles; one woman has long dreadlocks. They wear casual clothing like t-shirts, jeans, and sneakers.

\begin{table}[htbp]
\centering
\caption{Ablation study on training objectives. The models are trained on Wan2.1 generated videos and evaluated on the held-out test set. The metric is classification accuracy (\%). \textbf{Bold} indicates the best performance.}
\label{tab:bt}
\footnotesize
\setlength{\tabcolsep}{1.8mm}
\begin{tabular}{lcccccc}
\toprule
\textbf{Loss} & \textbf{[0, 0.2]} & \textbf{(0.2, 0.4]} & \textbf{(0.4, 0.6]} & \textbf{(0.6, 0.8]} & \textbf{(0.8, 1.0]} & \textbf{Avg} \\
\midrule
BT  & 73.50 & 77.25 & 78.25 & \textbf{82.50} & \textbf{87.75} & 79.85 \\
BCE & \textbf{77.00} & \textbf{78.50} & \textbf{79.75} & 82.00 & 83.00 & \textbf{80.05} \\
\bottomrule
\end{tabular}
\end{table}
\begin{table}[htbp]
\centering
\caption{Ablation study on the number of trainable DiT blocks. The metric is classification accuracy (\%). \textbf{Bold} indicates the best performance.}
\label{tab:rm_blocks}
\footnotesize
\setlength{\tabcolsep}{1.5mm}
\begin{tabular}{lcccccc}
\toprule
\textbf{Layer} & \textbf{[0, 0.2]} & \textbf{(0.2, 0.4]} & \textbf{(0.4, 0.6]} & \textbf{(0.6, 0.8]} & \textbf{(0.8, 1.0]} & \textbf{Avg} \\
\midrule
8         & 83.42 & 84.95 & 84.69 & 84.44 & 83.42 & 84.18 \\
16        & \textbf{84.95} & \textbf{85.46} & \textbf{85.71} & \textbf{86.73} & 84.69 & \textbf{85.51} \\
24        & 83.93 & 85.20 & \textbf{85.71} & 86.22 & \textbf{85.20} & 85.25 \\
32        & 80.36 & 83.67 & 84.95 & 85.97 & \textbf{85.20} & 84.03 \\
40 (Full) & 79.85 & 81.12 & 84.95 & 85.46 & 84.95 & 83.27 \\
\bottomrule
\end{tabular}
\end{table}

\begin{table*}[htbp]
\centering
\caption{Dataset analysis on PAVRMs, the model is based on Wan2.1.  The metric is average accuracy. Task is I2V and resoluition is 720P.}
\label{tab:rm_dataset}
\footnotesize
\begin{tabular}{llcccccc}
\toprule
\textbf{Train Set}&\textbf{Test Set}    & \textbf{[0, 0.2]} & \textbf{(0.2, 0.4]} & \textbf{(0.4, 0.6]} & \textbf{(0.6, 0.8]} & \textbf{(0.8, 1.0]} & \textbf{Avg} \\
\midrule
Veo3\&HunyuanVideo&Veo3\&HunyuanVideo& \textbf{77.00\%} & \textbf{86.00\%} & \textbf{89.00\%} & \textbf{89.00\%} & \textbf{87.00\%} & \textbf{85.60\%} \\
Wan2.1&Veo3\&HunyuanVideo& 70.00\% & 72.00\% & 73.00\% & 76.00\% & 81.00\% & 74.40\% \\
\bottomrule
\end{tabular}
\end{table*}

\paragraph{User Study.}
We ask each evaluator: For each question, two options represent videos generated from the title text using two different models. Select the option with the higher overall quality (greater text-video consistency, more natural motion, and no human deformities or physically implausible elements). There are 100 questions in total.
\begin{figure}
    \centering
    \includegraphics[width=0.8\linewidth]{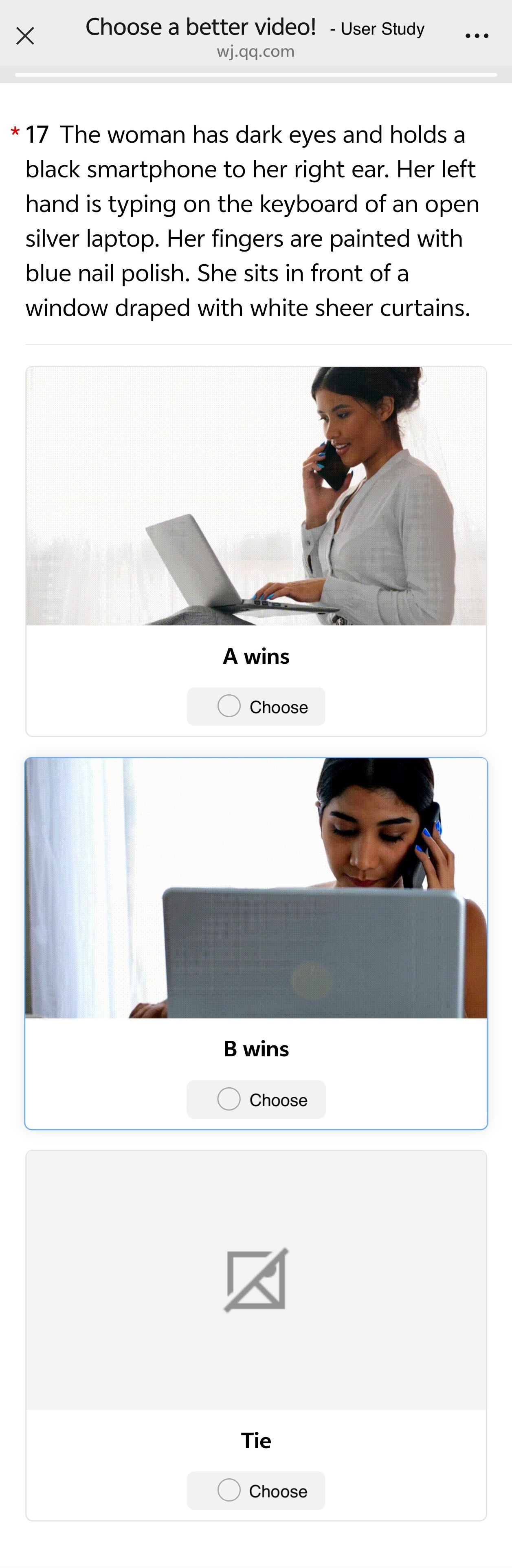}
    \vspace{-2mm}
    \caption{A case of user study page.}
    \label{fig:user_study_page}
    \vspace{-4mm}
\end{figure}
The page is shown in~\cref{fig:user_study_page}.

\section{More Experiments on Process-Aware Video Reward Models}

\label{sec:exp_more}
\subsection{The Influence of Training Loss}
To assess the robustness of our proposed method against different optimization objectives, we compare the standard binary cross-entropy (BCE) loss with the pairwise Bradley-Terry (BT) loss~\cite{bradley1952rank}.
Specifically, we construct preference pairs $(x_{\text{win}}, x_{\text{lose}})$ by randomly sampling a positive sample (label 1) as $x_{\text{win}}$ and a negative sample (label 0) as $x_{\text{lose}}$.
As shown in Table~\ref{tab:bt}, the average accuracy gap between the two objectives is marginal ($0.2\%$). Interestingly, we observe a trade-off across timesteps: BT loss performs better in high-noise regions ($t > 0.6$), while BCE demonstrates superior precision in the structure-forming and detailing stages ($t \le 0.6$).
Given that BCE achieves a slightly higher overall average accuracy and eliminates the computational overhead of pair construction, we adopt BCE as our default training objective.

\subsection{The Influence of the Number of DiT Blocks}
In our feasibility analysis, we observed that fixed DiT features are effective. Here, we investigate the impact of model depth when the DiT blocks are \textit{fully fine-tuned}. We vary the number of trainable DiT blocks (from the first 8 to the full 40 blocks) to determine the optimal capacity for the reward task.
The results in Table~\ref{tab:rm_blocks} reveal a non-monotonic trend, contradicting a simple scaling law. The performance peaks at 16 blocks (\textbf{85.51\%}) and subsequently degrades as more layers are added, with the full 40-block model performing worse than the 8-block baseline.
This suggests that the critical semantic information for assessing motion quality is concentrated in the early-to-middle layers of the network. Using the full generation backbone for the reward task is not only computationally expensive but potentially leads to optimization difficulties or overfitting to high-frequency generation details rather than high-level quality. Consequently, using the first 8 or 16 blocks offers the best trade-off between efficiency and accuracy.

\subsection{Cross-Model Generalization}
To evaluate the generalization capability of PAVRM, we extend our evaluation beyond the source domain. While our reward model is initialized and trained solely on data generated by Wan2.1, we test its performance on samples from two other state-of-the-art video generation models: HunyuanVideo~\cite{kong2024hunyuanvideo} and Veo3. The test sets for these models share the same annotation format but specifically focus on human structural deformities (e.g., limb distortions), a common challenge in video generation.

The results in~\cref{tab:rm_dataset} reveal two key insights regarding transferability and timestep sensitivity:

\noindent \textbf{Feasibility of Cross-Model Evaluation.}
First, PAVRM demonstrates strong zero-shot transferability. Despite being trained exclusively on Wan2.1 latents, it effectively identifies quality degradation in HunyuanVideo and Veo3. This suggests that the spatiotemporal features learned by the backbone VGM are not strictly model-specific but encode universal representations of motion and structure validity.

\noindent \textbf{Inverted Generalization across Timesteps.}
We observe a distinct behavior in performance distribution across diffusion timesteps ($t$) between in-domain and out-of-domain (OOD) settings:
\begin{itemize}
    \item \textbf{In-Domain (Wan2.1):} The model achieves higher performance in the middle and last trajectory ($t \in [0.2, 1]$). This aligns with the intuition that intermediate states balance signal and noise, containing the most critical information for motion formation.
    \item \textbf{Out-of-Domain (Hunyuan/Veo3):} Surprisingly, generalization is stronger in high-noise regions ($t \to 1$) compared to the near-data regions ($t \to 0$).
\end{itemize}

\noindent \textbf{Analysis.} We attribute this phenomenon to the nature of the denoising process. In the late stages of generation ($t \to 0$), the latents are dominated by model-specific high-frequency details and ``fingerprints'' (unique texture patterns or artifact types inherent to the specific generator architecture). A reward model trained on Wan2.1 overfits to these specific patterns, leading to poor transfer when evaluating clean latents from other models.
Conversely, at high noise levels ($t \to 1$), the latent representation is dominated by Gaussian noise and low-frequency structural layouts. The ``fingerprints'' of the specific generative model are less pronounced, while fundamental structural errors (such as severe human deformities) remain detectable as gross geometric inconsistencies. Consequently, the reward model relies on these universal structural cues rather than model-specific textures, resulting in superior generalization in high-noise regimes.

\end{document}